\documentclass[letterpaper, 10 pt, conference]{ieeeconf}
\IEEEoverridecommandlockouts
\usepackage[utf8]{inputenc} 
\usepackage[T1]{fontenc}    
\usepackage{hyperref}       
\usepackage{url}            
\usepackage{booktabs}       
\usepackage{amsfonts}       
\usepackage{nicefrac}       
\usepackage{microtype}      
\usepackage{xcolor}         
\usepackage{graphicx}       
\usepackage{multirow}       
\usepackage{colortbl}       
\usepackage{amsmath}
\usepackage{float}
\definecolor{bestblue}{RGB}{221,235,255}
\newcommand{\bestcell}[1]{\cellcolor{bestblue}\textbf{#1}}
\title{\LARGE \bf Learning Foresight without Explicit Trajectories for 3D Diffusion Policies}
\author{%
\authorblockN{Zhongbo Zhang\authorrefmark{1} \quad Zaibin Zhang\authorrefmark{1} \quad Yifan Wang \quad Changbo Yan \quad Lijun Wang\authorrefmark{2} \quad Huchuan Lu}%
\authorblockA{Dalian University of Technology}%
\thanks{\authorrefmark{1}Equal contribution. \authorrefmark{2}Corresponding author.}%
}

\begin{document}

\maketitle
\thispagestyle{empty}
\pagestyle{empty}

\begin{figure}[t]
\centering
\includegraphics[width=\columnwidth]{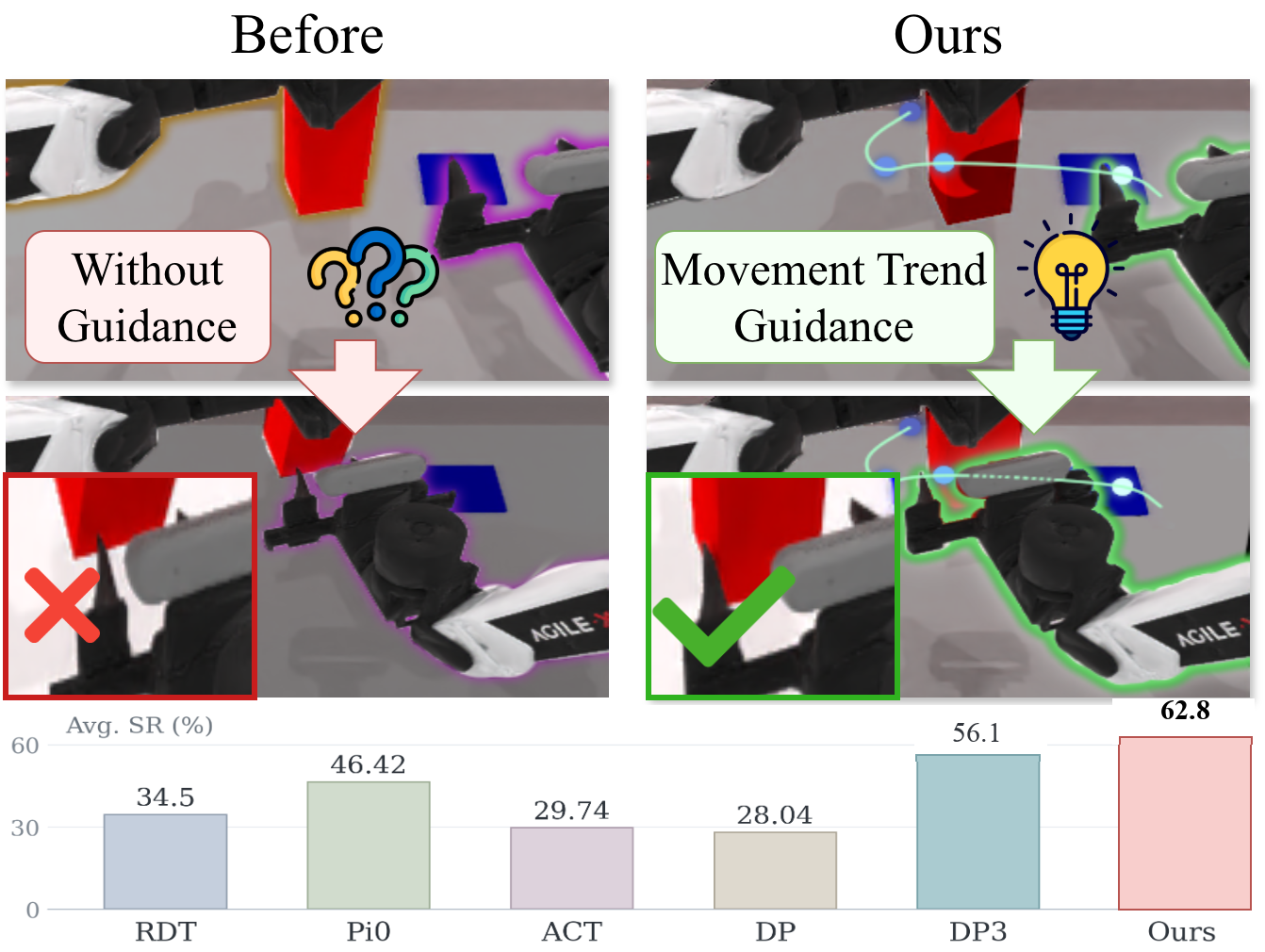}
\caption{Movement Trend Guidance for point-cloud diffusion policies. The latent trend provides soft geometric guidance that helps the policy organize future interaction while preserving observation-driven local action generation.}
\label{fig:teaser}
\end{figure}

\begin{abstract}
3D diffusion policies are strong at generating geometrically grounded actions from current observations, but successful manipulation requires not only knowing what motion is feasible now, but also anticipating where the interaction is heading.
Existing policies largely leave such foresight to emerge implicitly from action learning.
We introduce Movement Trend Guidance, a simple but effective way to provide this foresight without introducing an explicit plan. From a short observation history, the policy learns a compact latent representation of interaction evolution. During training, sparse future gripper states supervise this representation; at inference, only the latent is retained as future-oriented conditioning alongside the current observation. 
The latent provides global conditioning for action generation, while an additional gated FiLM branch is used only at the UNet bottleneck.
Despite adding only 3.52\% more parameters to DP3, our method preserves the original dense-action and receding-horizon formulation and consistently improves upon DP3 across RoboTwin2.0, LIBERO-40, and DexArt. It reaches 62.8\% vs. 56.1\% in 50-task RoboTwin2.0 mixed training, 71.93\% vs. 37.08\% on LIBERO-40, and 72.0\% vs. 49.0\% on five real-robot tasks. These results show that a diffusion policy can benefit substantially from knowing where an interaction is heading, without being told exactly where to move. Code is available at \url{https://github.com/zhangzhongbo2213/movement-trend-guidance}.

\end{abstract}

\section{Introduction}
3D diffusion policies~\cite{ze2024dp3,gervet2023act3d,hong2026r3d} provide a compelling formulation for robot manipulation: they map geometric observations directly to dense action sequences. This makes them well suited to precise control~\cite{ze2024dp3}. Yet manipulation is not determined solely by the motion that is feasible at the current instant. Before grasping, inserting, transferring an object, or operating an articulated mechanism, the robot must also understand \textit{how the interaction is unfolding and what it is progressing toward}.

This creates a distinction between two kinds of information. Current geometry answers what action is appropriate now. Nonetheless, successful multi-step behavior additionally requires a notion of where the interaction is heading: which contact should occur next, or toward which configuration multiple effectors should coordinate. Standard diffusion policies do not explicitly separate these roles. Both local control and longer-range interaction structure must be recovered through the same action-learning objective.

A straightforward way to expose future structure is to predict waypoints, keyframes~\cite{li2026causal,intelligence2026pi}, or trajectories~\cite{janner2022planning}. However, doing so asks the predictor for more than the controller may actually need. A trajectory specifies where the robot should move, whereas action generation may only require a coarse indication of how the interaction is developing. Once a predicted future is converted into an intermediate control target, prediction errors can also constrain subsequent behavior, even when new observations suggest that the motion should be corrected.

We therefore consider a simpler use of future prediction: use it to teach the policy what future-relevant information to represent, rather than what future motion to execute. Our method, Movement Trend Guidance, predicts a compact latent representation from a short history of point clouds and robot states. During training, an auxiliary objective requires this latent to recover sparse future gripper states. During inference, only the latent representation, rather than the decoded future states, is retained as future-oriented conditioning for the action policy.

The architecture follows the same principle of minimal intervention. The trend latent enters the standard global-conditioning pathway, while only its additional gated FiLM modulation is restricted to the compressed bottleneck of the diffusion UNet. This small change adds no separate planning module and preserves dense action prediction and receding-horizon execution.

Despite its simplicity, this design produces consistent gains across different manipulation settings. On 50-task RoboTwin2.0 mixed training, it improves the matched DP3 baseline from 56.1\% to 62.8\%; on LIBERO-40, from 37.08\% to 71.93\%; and on five real-robot tasks, from 49.0\% to 72.0\%. Controlled studies further show that future supervision improves the learned latent beyond additional representation capacity alone, while directly conditioning on explicit future points is less effective than using future prediction to shape the latent representation.

Our contributions are summarized as follows:
\begin{itemize}
    \item We introduce a simple formulation of foresight for diffusion control: the policy learns not only what can be done at the current step, but also where the interaction is heading, without requiring an explicit planning target.
    
    \item We present a lightweight realization in which future states supervise a compact latent representation. The latent enters the standard global-conditioning pathway, while only its additional gated FiLM modulation is restricted to the UNet bottleneck.
    
    \item We provide broad empirical validation across RoboTwin~2.0, LIBERO-40, DexArt, and real-world robotic manipulation. The proposed method consistently improves 3D diffusion control while preserving the underlying diffusion-policy architecture.
\end{itemize}

\section{Related Work}

\subsection{Robot Manipulation}

Sequence and action-chunking policies such as ACT~\cite{zhao2023act}, BeT~\cite{shafiullah2022bet}, and VQ-BeT~\cite{lee2024vqbet} improve imitation learning by modeling extended-horizon actions, while vision-language-action models extend this paradigm to multi-task policy learning~\cite{brohan2023rt2,kim2024openvla,black2024pi0,o2024open}.
Diffusion Policy~\cite{chi2023diffusionpolicy} shows conditional diffusion captures multimodal action distributions and generates coherent action chunks, while later work extends diffusion control to broader settings and faster inference~\cite{janner2022planning,lu2024manicm}.
These methods provide strong dense action generators, but future interaction structure typically emerges implicitly from the action-learning objective.
Our work preserves dense diffusion-based action generation while introducing a future-supervised latent representation of movement trend.

\subsection{3D Imitation Learning Policies}

3D observations aid manipulation by exposing geometry, spatial relations, and viewpoint-robust structure~\cite{qi2017pointnet,qi2017pointnet++,zhao2021point,shridhar2022peract}.
Prior 3D imitation learning methods use voxelized observations, neural fields, multiview features, or adaptive-resolution representations to predict actions or keyframes~\cite{ze2023gnfactor,goyal2023rvt,goyal2024rvt2,gervet2023act3d,ke2024diffuseractor}.
Many methods make intermediate structure explicit through keyframe pose prediction or prediction-and-planning, but rely on pose-level targets, explicit tracking, or planning interfaces distinct from direct dense action generation.
DP3~\cite{ze2024dp3} uses point-cloud observations with direct action diffusion, achieving strong sample efficiency and real-robot transfer.
We follow DP3's point-cloud diffusion formulation, using sparse future-state supervision to shape a compact movement-trend latent jointly with action generation.

\subsection{Future-Guided Policy Learning}

Several policy-learning frameworks use future goals, subgoals, keyframes, or trajectories to provide longer-horizon structure for manipulation.
Recent diffusion-based approaches also introduce explicit guidance; HDP~\cite{wang2024hdp} leverages contact structure, while DTP~\cite{fan2025dtp} generates task-relevant 2D trajectories to guide manipulation.
FLARE~\cite{zheng2025flare} and ForeDiffusion~\cite{xie2026forediffusion} introduce foresight through predictive representations of future visual observations. FLARE aligns policy features with future-observation latents, while ForeDiffusion predicts and injects a future-view representation. In contrast, our method operates directly in 3D space. From point-cloud and robot-state history, sparse task-space gripper states at multiple future offsets serve only as training-time supervision for a compact movement-trend latent. The decoded future states are discarded at inference, retaining only the latent as future-oriented conditioning alongside the current observation, with direct modulation restricted to the UNet bottleneck. Thus, our method differs in the future information modeled and its coupling to control, providing task-space foresight without future-view prediction or explicit future trajectories.

\subsection{Conditional Modulation}

Classifier-free guidance~\cite{ho2022cfg} controls conditioning strength in generative models, while ControlNet~\cite{zhang2023controlnet} introduces pathways for external conditions.
Feature-wise modulation provides a lightweight mechanism for conditioning intermediate representations.
These approaches motivate careful control of how auxiliary information interacts with a generative backbone.
Here, the movement-trend latent provides a compact future-oriented condition rather than a dense control target.
The trend latent enters the standard global-conditioning pathway, while its gated FiLM modulation is restricted to the UNet bottleneck.
This allows future-supervised information to influence broader action-sequence structure while preserving the original pathway for observation-driven local refinement.
\section{Method}

\begin{figure*}[t]
\centering
\includegraphics[width=\textwidth,trim=22 670 22 32,clip]{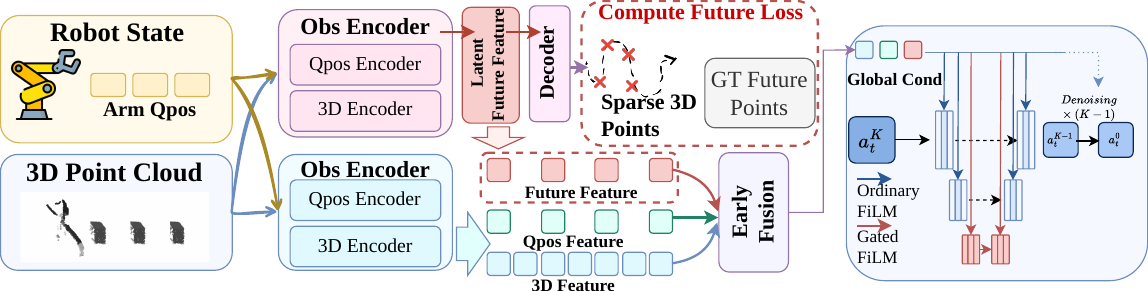}
\caption{Overview of Movement Trend Guidance. The sparse-point symbols denote the Cartesian components of future gripper targets; each target also contains the corresponding gripper state defined in Eq.~\eqref{eq}. During training, the auxiliary decoder shapes the latent representation, while its outputs are discarded at inference and never supplied to the action policy. The latent enters the standard global-conditioning pathway, with an additional gated FiLM branch applied only at the UNet bottleneck.}
\label{fig}
\end{figure*}

This section presents movement-trend guidance for 3D diffusion policies. A standard observation-to-action policy must infer scene geometry, interaction intent, and precise local controls jointly. We instead expose a hidden latent from observation history that compresses future interaction information and provides soft geometric guidance while preserving flexible dense action generation without a decoded trajectory.

\subsection{Problem Formulation}

At time $t$, the policy receives an observation window of length $N$,
\begin{equation}
\mathcal{O}t = {o{t-N+1}, \ldots, o_t},
\end{equation}
where each observation $o_i=(P_i,q_i)$ consists of a point cloud $P_i$ and robot state $q_i$. As in DP3, the length-$H$ prediction is aligned with the observation window and extends into the future:
\begin{equation}
\mathbf{A}t={a{t-N+1},\ldots,a_t,\ldots,a_{t-N+H}}.
\end{equation}
The current action $a_t$ is therefore at zero-based index $N-1$. Following DP3's receding-horizon setting, we set $N=3, H=8$ and execute $n_a=6$ actions from this index before re-observing.

We represent future movement information with a compact latent rather than an explicit spatial trajectory. Let $F_\phi$ denote a lightweight future-information encoder mapping the observation window to
\begin{equation}
z_t = F_\phi(\mathcal{O}_t) \in \mathbb{R}^{d_z}, \qquad d_z=256.
\end{equation}
The vector $z_t$ is the latent preceding the auxiliary output head, which provides future gripper-target supervision during training but whose decoded values are never used as policy inputs. Thus, $z_t$ represents a compressed movement trend rather than future coordinates.
The action policy predicts action sequences from the observation and future trend latent:
\begin{equation}
\hat{\mathbf{A}}t=\Pi\theta(\mathcal{O}_t,z_t).
\end{equation}
This factorization exposes future-dependent information without making it an intermediate control target. The latent summarizes interaction evolution, while the point-cloud encoder and action denoiser remain responsible for geometry and local control.

\subsection{Latent Movement-Trend Representation}

Each observation frame is encoded by a DP3-style point-cloud and state encoder,
\begin{equation}
e_i=E_{\mathrm{trend}}(P_i,q_i), \qquad i\in{t-N+1,\ldots,t}.
\end{equation}
Before encoding, point-cloud coordinates and robot states are scaled to $[-1,1]$ using per-dimension limits estimated from training data. This input normalization is distinct from future-target normalization.
The encoded features and robot states are concatenated temporally and passed through a two-hidden-layer MLP:
\begin{equation}
z_t = F_\phi\left([e_{t-N+1};\ldots;e_t], [q_{t-N+1};\ldots;q_t]\right).
\end{equation}
The MLP hidden width is 256, and the latent is taken from the hidden layer immediately before the auxiliary future gripper-target head. The policy therefore receives the future-relevant representation, while the auxiliary head provides a compact training signal rather than a point-by-point tracking trajectory.
The trend latent is first projected by a small MLP,
\begin{equation}
u_t=\psi(z_t).
\end{equation}
For each observation frame, point-cloud and robot-state features are encoded independently and fused with the shared trend feature:
\begin{equation}
f_i^{P}=E_P(P_i), \qquad
f_i^{q}=E_q(q_i), \qquad
h_i=F_{\mathrm{fuse}}([f_i^{P};f_i^{q};u_t]).
\end{equation}
The same $u_t$ is shared across all $N$ observation frames, and the resulting features form the global diffusion condition,
\begin{equation}
c_t=[h_{t-N+1};\ldots;h_t].
\end{equation}

In the 50-task RoboTwin2.0 experiments, both $\mathit{DP3}^{}$ and $\mathit{Ours}^{}$ use the same learned 64-dimensional task embedding, projected into policy features to distinguish task identities. $\mathit{DP3}^{}$ otherwise follows vanilla DP3, whereas $\mathit{Ours}^{}$ additionally uses the movement-trend pathway and bottleneck modulation. Thus, task conditioning is matched, and controlled ablations retain the same embedding.

The latent trend acts as soft geometric guidance rather than a waypoint. The policy combines it with current point-cloud and robot-state observations to infer and execute local motions.

\subsection{Action Diffusion with Bottleneck-Gated Guidance}

Given a ground-truth action block $\mathbf{A}t$, the forward diffusion process~\cite{ho2020denoising} produces
\begin{equation}
x\tau=\sqrt{\bar{\alpha}\tau},\mathbf{A}t
+\sqrt{1-\bar{\alpha}\tau},\epsilon, \qquad \epsilon\sim\mathcal{N}(0,I).
\end{equation}
The denoising network predicts the clean action sequence,
\begin{equation}
\hat{\mathbf{A}}t=D\theta(x\tau,\tau,c_t).
\end{equation}
We use the sample-prediction objective,
\begin{equation}
\mathcal{L}{\mathrm{diff}}=
\mathbb{E}{\mathbf{A}t,\epsilon,\tau}
\left[|D\theta(x_\tau,\tau,c_t)-\mathbf{A}_t|_2^2\right].
\end{equation}

The latent-conditioned representation is available through the standard global-condition pathway. To regulate its direct influence on the denoising backbone, we additionally apply bottleneck-gated FiLM. Let $x$ be the bottleneck feature and $c$ the concatenated diffusion-time and observation condition:
\begin{equation}
\alpha=W_\alpha c, \qquad \beta=W_\beta c, \qquad
\gamma=\sigma(W_\gamma c+b_\gamma),
\end{equation}
\begin{equation}
\tilde{x}=(1+\alpha)\odot x+\beta, \qquad
y=x+\gamma\odot(\tilde{x}-x).
\end{equation}
The scale and bias projections are initialized to zero, making the additional branch an exact identity mapping initially, independent of the gate value. A negative gate bias further limits modulation as the projections begin to learn. The gated branch is restricted to the UNet bottleneck, while down-sampling and up-sampling blocks retain the ordinary latent-conditioned pathway. Thus, the latent remains available through global conditioning, with only direct additional modulation confined to the bottleneck.

\subsection{End-to-End Training and Inference}

We jointly optimize the trend encoder, latent predictor, latent embedding, observation encoder, and action diffusion model with a single end-to-end objective. The auxiliary decoder maps the latent to $K=4$ future gripper targets,
\begin{equation}
\hat{\mathbf{P}}t=G\omega(z_t)
\in\mathbb{R}^{K\times 2\times 4},
\qquad \Delta={5,10,15,20}.
\end{equation}
For offset $\delta_k$ and gripper slot $g$, the target
$\mathbf{p}^{(g)}_{t+\delta_k}\in\mathbb{R}^{4}$ contains Cartesian end-effector position $(x,y,z)$ and one scalar gripper state. The two slots correspond to the left and right grippers in bimanual tasks; for single-arm LIBERO tasks, the target is duplicated across both slots to preserve the tensor interface.

Offsets index the trajectory supplied by each benchmark. In RoboTwin2.0, stride-4 sampling changes only training-window anchors, so offsets remain measured in original demonstration frames. In LIBERO-40, they index the stride-4 materialized trajectory. Targets use each benchmark's stored Cartesian frame and enter the loss directly without separate future-target normalization. Offsets beyond a demonstration are clipped to its final frame. We use the elementwise mean-squared error
\begin{equation}
\label{eq}
\mathcal{L}{\mathrm{future}}=
\frac{1}{8K}\sum{k=1}^{K}\sum_{g=1}^{2}
\left|\hat{\mathbf{p}}^{(g)}{t+\delta_k}
-\mathbf{p}^{(g)}{t+\delta_k}\right|2^2,
\end{equation}
and optimize
\begin{equation}
\mathcal{L}=\mathcal{L}{\mathrm{diff}}+
\lambda_{\mathrm{future}}\mathcal{L}{\mathrm{future}},
\qquad \lambda{\mathrm{future}}=1.0.
\end{equation}
The decoder produces $4\times2\times4=32$ scalars as a compact geometric training signal, not a waypoint sequence or tracking target. The parameter-matched ablation sets $\lambda_{\mathrm{future}}=0$ while retaining the latent architecture.

During inference, the policy denoises Gaussian noise under $c_t$ and executes
\begin{equation}
\mathbf{a}^{\mathrm{exec}}_t=
\hat{\mathbf{A}}t[N-1+n_a]
={\hat a_t,\ldots,\hat a{t+n_a-1}}.
\end{equation}
After each executed subsequence, the robot obtains a new calibrated point-cloud observation, recomputes $z_t$, and replans. The latent is refreshed every receding-horizon cycle and never directly converted into a command or tracking path.
\section{Experiments}

We evaluate the proposed movement-trend-guided DP3 on simulation benchmarks and real-robot manipulation tasks.
We evaluate on RoboTwin2.0~\cite{chen2025robotwin}, a large-scale bimanual manipulation benchmark with 50 diverse tasks.
We further evaluate and analyze our method in the distinct environments of LIBERO-40~\cite{liu2023libero} and DexArt~\cite{bao2023dexart} to assess its generality.
Finally, we evaluate the method on five SO101 real-robot manipulation tasks.

\subsection{Experimental Setup}
\label{sec:setup}

We evaluate on RoboTwin2.0, LIBERO-40, DexArt, and five SO101 real-robot tasks.
For RoboTwin2.0, each task contains 50 demonstrations and is evaluated over 100 episodes. We first reproduce the official RoboTwin2.0 setting. We also train one policy jointly on all 50 tasks to assess multi-task capability. This mixed-training setting uses 50 demonstrations per task. We retain every fourth frame as a training-window anchor while preserving consecutive frames within each sampled window. The policy is trained for 3000 epochs, matching the original training schedule. Evaluation strictly follows the official protocol for a fair comparison.

For LIBERO-40, each task contains 50 demonstrations. We use trajectories materialized with stride-4 temporal subsampling and train for 1000 epochs. Checkpoints saved every 100 epochs are evaluated with rollout seeds 7, 17, and 27, using 50 episodes per task and seed. DexArt uses 100 demonstrations per task and 3000 training epochs. Policies are evaluated every 200 epochs with 20 episodes per task, and the reported score averages the best five evaluation checkpoints. This benchmark-specific aggregation differs from the fixed epoch-1000 LIBERO report. Each real-robot task uses 50 demonstrations and 20 evaluation rollouts.
Our DP3 reproductions and movement-trend variants use AdamW with a learning rate of $1.0\times10^{-4}$.
We report success rate as the primary metric.

Our method contains 271.67M parameters, adding 9.23M (+3.52\%) over the 262.43M-parameter DP3 baseline. On an NVIDIA RTX 4090 with batch size 1 and 10 DDIM denoising steps, this increases mean inference latency only from 50.28 ms to 50.90 ms (+1.23\%).

\subsection{Main Results on RoboTwin2.0}
\label{sec:main_results}

Table~\ref{tab:robotwin_category} retains the original RoboTwin2.0 category-level comparison and appends the new 50-task mixed-training results as the final two rows. Each entry is a success rate (\%), averaged over the tasks in the corresponding category.

\begin{table*}[t]
\caption{RoboTwin2.0 category-level success rates (\%). The original comparison rows are retained. The starred rows are the newly supplied 50-task mixed-training results for $\mathit{DP3}^{*}$ and $\mathit{Ours}^{*}$, using 50 demonstrations per task and 100 evaluation episodes per task. Both starred policies receive the same learned 64-dimensional task embedding; $\mathit{DP3}^{*}$ otherwise retains vanilla DP3 conditioning, while $\mathit{Ours}^{*}$ additionally includes the proposed trend pathway and bottleneck modulation. Values are averaged over the tasks in each category; the final column is the unweighted average over all 50 tasks.}
\label{tab:robotwin_category}
\centering
\resizebox{\textwidth}{!}{%
\begin{tabular}{@{}lcccccccc@{}}
\toprule
\textbf{Method} & Single-Arm & Articulated & Bimanual & Pick \& & Precise & Multi-Object & Tool Use & Overall \\
 & Handling & Interaction & Handover & Transport & Placement & Stacking & \& Precision & Average \\
\midrule
RDT             & 67.4 & 51.9 & 67.5 & 22.0 & 20.2 & 37.5 & 17.0 & 34.5 \\
$\pi_0$         & 80.0 & 61.3 & 71.5 & 42.3 & 32.1 & \bestcell{54.0} & 18.0 & 46.4 \\
ACT             & 70.1 & 38.4 & 63.5 & 17.3 & 14.5 & 38.8 & 12.2 & 29.7 \\
DP              & 59.3 & 32.0 & 31.5 & 22.8 & 19.6 & 32.8 & 10.8 & 28.0 \\
DP3             & 84.9 & 71.3 & 85.0 & 54.4 & 45.5 & 41.3 & 25.2 & 55.2 \\
Ours            & \bestcell{88.9} & 78.0 & \bestcell{95.0} & 61.9 & 52.0 & 51.0 & 29.0 & 61.7 \\
\midrule
$\mathit{DP3}^{*}$  & 79.9 & 72.4 & 85.5 & 55.6 & 49.3 & 37.8 & 27.2 & 56.1 \\
$\mathit{Ours}^{*}$ & 87.7 & \bestcell{80.0} & 94.0 & \bestcell{62.9} & \bestcell{54.8} & 48.3 & \bestcell{29.8} & \bestcell{62.8} \\
\bottomrule
\end{tabular}
}
\end{table*}

$\mathit{Ours}^{*}$ reaches 62.8\%, a +6.7-point gain over $\mathit{DP3}^{*}$, and improves on 38 of the 50 tasks; the remaining tasks show broadly comparable performance overall. Because the two starred systems use the same task embedding, this comparison directly evaluates the addition of the trend pathway and bottleneck modulation. Component-level attribution is further examined in the controlled ablations in Section~\ref{sec:ablation}, where the task embedding is held fixed. The largest task-level gains occur on contact-rich and multi-stage tasks such as \texttt{rotate\_QRcode} (+37), \texttt{move\_playingcard\_away} (+21), \texttt{move\_pillbottle\_pad} (+22), and \texttt{place\_shoe} (+26).

\subsection{Evaluation Across Environments}
\label{sec:cross_environment}

We evaluate the method in two distinct simulated environments, LIBERO-40 and DexArt, using their respective official task definitions and evaluation protocols.

\subsubsection{LIBERO-40 Point-Cloud Evaluation}
\label{sec:libero_results}

\begin{figure}[t]
\centering
\includegraphics[width=\columnwidth]{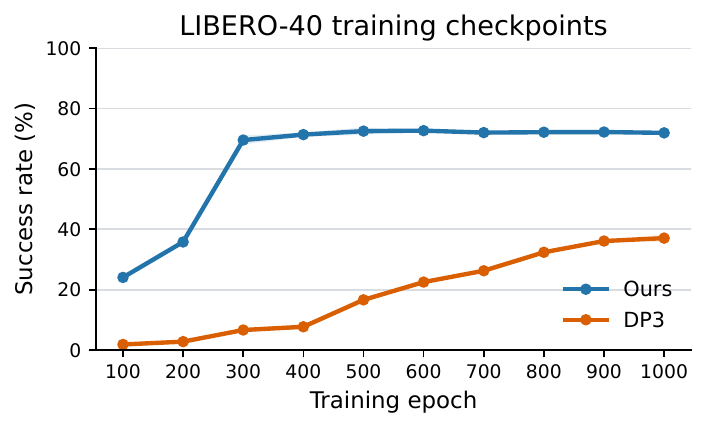}
\caption{LIBERO-40 point-cloud evaluation. Overall success rate for checkpoints saved every 100 epochs from epoch 100 to 1000.}
\label{fig:libero_latent_dp3}
\end{figure}

To compare our method directly with the point-cloud baseline, we evaluate both policies on the 40-task LIBERO suite.
The four suites (LIBERO-Spatial, LIBERO-Object, LIBERO-Goal, and LIBERO-10) use the same 50 training demonstrations per task, the same calibrated point-cloud preprocessing, and the same evaluation initialization protocol.
For each checkpoint saved every 100 epochs from epoch 100 to 1000, evaluation uses rollout seeds 7, 17, and 27 with 50 episodes per task and seed, totaling 6,000 episodes per checkpoint.
Figure~\ref{fig:libero_latent_dp3} shows the overall success rate across these saved checkpoints, while Table~\ref{tab:libero_latent_dp3} reports the suite-level and overall epoch-1000 results as mean $\pm$ sample standard deviation across the three seed-level success rates.

\begin{table*}[t]
\caption{LIBERO-40 point-cloud evaluation at epoch 1000. Values are success rates (\%) reported as mean $\pm$ sample standard deviation across rollout seeds 7, 17, and 27; Fig.~\ref{fig:libero_latent_dp3} uses the same aggregation for checkpoints saved every 100 epochs. Each suite contains 10 tasks and 1,500 evaluation episodes. Overall contains all 40 tasks and 6,000 episodes.}
\label{tab:libero_latent_dp3}
\centering
\small
\setlength{\tabcolsep}{10pt}
\begin{tabular}{@{}lccccc@{}}
\toprule
Suite & Tasks & DP3 & Ours & Improvement & Episodes \\
\midrule
LIBERO-Spatial & 10 & $26.40\pm1.31$ & \bestcell{$58.20\pm0.72$} & $+31.80\pm1.22$ & 1,500 \\
LIBERO-Object  & 10 & $67.53\pm1.14$ & \bestcell{$93.53\pm0.70$} & $+26.00\pm1.44$ & 1,500 \\
LIBERO-Goal    & 10 & $8.73\pm0.76$  & \bestcell{$77.53\pm2.32$} & $+68.80\pm2.62$ & 1,500 \\
LIBERO-10      & 10 & $45.67\pm1.80$ & \bestcell{$58.47\pm0.64$} & $+12.80\pm2.42$ & 1,500 \\
\midrule
\textbf{Overall} & 40 & $37.08\pm1.21$ & \bestcell{$71.93\pm0.49$} & $+34.85\pm1.30$ & 6,000 \\
\bottomrule
\end{tabular}
\end{table*}

Our method substantially outperforms DP3 on all four suites, reaching $71.93\%\pm0.49\%$ overall at epoch 1000 and a $+34.85$-point gain over DP3.
The corresponding LIBERO guidance ablations are analyzed in Section~\ref{sec:abl_libero}.
The largest absolute gain is on LIBERO-Goal.
The narrow seed deviations indicate that the advantage is consistent across the three evaluation seeds rather than being driven by one rollout stream.

\subsubsection{DexArt}
\label{sec:cross_benchmark}

We further evaluate the method on four DexArt tasks.
DexArt features dexterous articulated object manipulation and differs from RoboTwin2.0 in robot morphology, task structure, and object categories.

\begin{table}[t]
\caption{DexArt success rates (\%) with representative baselines from the DexArt evaluation setting. Our result averages the best five checkpoints under the evaluation schedule described in Section~\ref{sec:setup}. Best results are bolded and shaded in blue.}
\label{tab:cross_benchmark_results}
\centering
\scriptsize
\setlength{\tabcolsep}{2.5pt}
\begin{tabular}{@{}lccccc@{}}
\toprule
\multicolumn{6}{c}{DexArt} \\
\midrule
Method & faucet & laptop & toilet & bucket & Avg. \\
\midrule
IBC                  & 7  & 3  & 0  & 14 & 6.0  \\
BCRNN                & 1  & 3  & 0  & 5  & 2.3  \\
H$^3$DP              & 34 & 81 & 70 & 28 & 53.3 \\
FreqPolicy           & 30 & 85 & \bestcell{77} & 25 & 54.3 \\
DP                   & 23 & 69 & 58 & 20 & 42.5 \\
SimpleDP3            & 26 & 79 & 63 & 22 & 47.5 \\
DP3                  & 36 & 77 & 70 & 25 & 52.0 \\
VITA                 & \bestcell{39} & 82 & 72 & 28 & 55.3 \\
Ours                 & \bestcell{41} & \bestcell{87} & 76 & \bestcell{33} & \bestcell{59.25} \\
\bottomrule
\end{tabular}
\end{table}

Our method reaches 59.25\% average success on DexArt, improving over DP3 by 7.25 points and achieving the highest average among the compared methods.

\subsection{Real-Robot Experiments}
\label{sec:real_robot}

We further conduct real-robot experiments on the SO101 dual-arm platform on five manipulation tasks: push cube, stack bowls, stack cubes, lift basket, and handover bottle.
The policy receives point clouds reconstructed from Intel RealSense RGB-D observations: synchronized stereo depth is deprojected using the calibrated camera intrinsics and transformed into the robot coordinate frame with camera-to-robot extrinsics.
Each task uses 50 demonstrations and is evaluated over 20 rollouts.
Table~\ref{tab:real_robot_results} reports the resulting success rates.
Our method reaches 72.0\% average success, compared with 49.0\% for DP3 and 43.0\% for SimpleDP3.

\begin{table}[t]
\caption{Real-robot success rates (\%) on five tasks. Best results are bolded and shaded.}
\vspace{-0.5em}
\label{tab:real_robot_results}
\centering
\scriptsize
\setlength{\tabcolsep}{2.5pt}
\renewcommand{\arraystretch}{0.88}
\resizebox{\linewidth}{!}{%
\begin{tabular}{@{}lcccccc@{}}
\toprule
Method & Push Cube & Stack Bowls & Stack Cubes & Lift Basket & Handover Bottle & Avg. \\
\midrule
DP        & 40 & 15 & 60 & 20 & 30 & 33.0 \\
SimpleDP3 & 40 & 25 & 70 & 30 & 50 & 43.0 \\
DP3       & 50 & 30 & 75 & 35 & 55 & 49.0 \\
Ours       & \bestcell{75} & \bestcell{75} & \bestcell{85} & \bestcell{45} & \bestcell{80} & \bestcell{72.0} \\
\bottomrule
\end{tabular}
}
\vspace{-0.4em}
\end{table}

\subsection{Failure Cases}
\label{sec:failure_cases}

Some tasks show only modest improvements over DP3. Our failure analysis identifies a common geometric cause: the robot arm can occlude most of the manipulated object, leaving an incomplete point cloud. The missing geometry affects both policies and can also perturb the movement-trend estimate used by our method. The effect is most visible in tasks that require precise contact, grasp alignment, or placement. In these cases, a trend estimate based on partial observations may provide only limited additional information. This behavior is consistent with our design: the movement-trend latent provides soft guidance, so the policy must still combine it with the current point-cloud and robot-state observations to infer local actions.

Our real-robot experiments demonstrate transfer in the evaluated SO101 setting; broader validation across diverse objects, scenes, and perturbations remains an important direction for future work.

\section{Ablation Studies}
\label{sec:ablation}

\begin{figure*}[t]
\centering
\includegraphics[width=\textwidth]{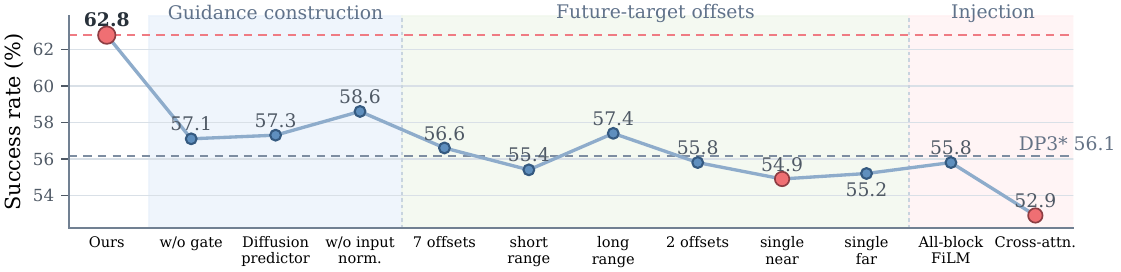}
\caption{Ablation dimensions under the 50-task mixed-training setting. We vary guidance construction, future-target offsets, and the scope of bottleneck-gated injection; every point is obtained from a shared policy trained jointly on all 50 RoboTwin2.0 tasks. The observation-history length remains fixed.}
\label{fig:ablation}
\end{figure*}

We ablate key design choices on RoboTwin2.0.
Unless otherwise noted, the DP3-based ablation variants are trained jointly on the complete 50-task suite using the same mixed-task sampling, demonstration budget, and evaluation protocol as the main experiment. Each reported value therefore evaluates one shared-policy run rather than an independently trained task-specific policy. The ACT experiment in Section~\ref{sec:abl_act} is a separate six-task validation.
All variants use the default future-target offsets unless otherwise specified.
Our implementation uses a learned latent trend representation rather than an explicit sparse-point sequence. The ablations therefore focus on the latent guidance path, its training signal, its future-target offsets, and the architecture used to inject it into the diffusion model.
Figure~\ref{fig:ablation} summarizes the ablation dimensions under this shared-policy protocol.

\subsection{Guidance Construction and Training}
\label{sec:abl_framework}

This group validates the latent guidance path, bottleneck-gated modulation, and training objective while keeping joint training over all 50 RoboTwin2.0 tasks fixed.
Removing gated FiLM leaves the latent trend available through early fusion and reduces the model to a DP3-style diffusion policy with an enriched global condition.
The comparison tests whether the latent is useful merely as extra representation capacity or whether bottleneck regulation provides a distinct benefit.
Additional parameter-matched variants replace the default MLP trend predictor with a diffusion predictor or remove input normalization while preserving the shared 50-task protocol. The latter bypasses limits normalization for the point-cloud and robot-state inputs of the trend predictor; action normalization and the unnormalized future targets are unchanged.
These comparisons test whether the observed gain depends on the specific predictor and preprocessing choices.

\subsection{LIBERO-40 Guidance Ablations}
\label{sec:abl_libero}

To test whether the same design choices transfer beyond RoboTwin2.0, we repeat the comparison on LIBERO-40 under a fixed point-cloud protocol.
Each variant uses 50 demonstrations per task, stride-4 temporal subsampling, and 1000 training epochs. Evaluation at epoch 1000 uses rollout seeds 7, 17, and 27 with 50 episodes per task and seed.
Table~\ref{tab:libero_ablations} reports the four-suite success rate as the mean $\pm$ sample standard deviation across the three seeds.
The complete model achieves the highest result, $71.93\pm0.49\%$. The parameter-matched latent variant without future supervision reaches $44.87\pm0.33\%$, only $+7.79$ points above DP3. Adding future supervision yields a further $+27.06$ points. The complete model also exceeds explicit future-point conditioning by $+6.65$ points. These results show that future-supervised representation learning, rather than additional latent capacity alone, drives most of the improvement.

\begin{table}[t]
\caption{Compact LIBERO-40 ablation results at epoch 1000. Values are four-suite success rates reported as mean $\pm$ sample standard deviation across rollout seeds 7, 17, and 27.}
\label{tab:libero_ablations}
\centering
\scriptsize
\setlength{\tabcolsep}{4pt}
\begin{tabular}{@{}lp{2.30cm}c@{}}
\toprule
Method & Future guidance & SR (\%) \\
\midrule
DP3 & None & $37.08\pm1.21$ \\
Ours & Latent trend + future loss & $\mathbf{71.93\pm0.49}$ \\
Parameter-Matched Latent & Same latent, no future loss & $44.87\pm0.33$ \\
Explicit Future-Point Conditioning & Points at 5, 10, 15, 20 embedded in condition & $65.28\pm0.19$ \\
\bottomrule
\end{tabular}
\end{table}

\subsection{Future-Target Offset Configuration}
\label{sec:abl_offset}

This group varies the future offsets used by the auxiliary supervision while keeping the three-frame observation history fixed in the same 50-task mixed-training run. Besides the default $\{5,10,15,20\}$, we evaluate seven offsets $\{5,10,15,20,25,30,35\}$, short-range offsets $\{2,4,6,8\}$, long-range offsets $\{20,30,40,50\}$, two offsets $\{5,15\}$, a single near offset $\{5\}$, and a single far offset $\{30\}$. These values are original demonstration-frame offsets; stride-4 sampling changes only the training-window anchors.
Short-range targets can be redundant with the observation window, whereas overly long ranges can dilute the compact representation with details that are better handled by the action diffusion model.
The default offsets balance near-term contact relevance and longer-horizon movement direction.
These targets shape the latent during training; they are not intermediate coordinates that the executed actions must reach.

\subsection{Bottleneck-Gated FiLM Injection}
\label{sec:abl_injection}

This group examines how latent movement-trend guidance enters the denoising backbone, with every injection variant trained jointly on all 50 tasks.
The full method applies gated FiLM only at the UNet bottleneck, while the standard conditional pathway carries the latent-conditioned global representation through the network.
Applying gated FiLM at every UNet block lets the high-level latent directly perturb layers responsible for local action refinement, while cross-attention creates an even stronger coupling between denoising features and the trend condition.
Bottleneck-only gated FiLM reaches 62.8\%, compared with 55.8\% for all-block gated FiLM and 52.9\% for cross-attention under the same protocol.
These experiments test the conservative injection hypothesis: latent movement-trend guidance should shape global action structure without dominating local denoising.
All controlled variants use the same 64-dimensional task embedding so that one shared latent trend encoder can distinguish task-specific interaction semantics while being trained on the full 50-task mixture.

\subsection{Transfer Across Policy Architectures}
\label{sec:abl_act}

To test whether movement-trend conditioning is specific to the DP3 diffusion UNet, we conduct a separate validation with ACT, a transformer-based imitation policy, on the six RoboTwin2.0 tasks listed in Table~\ref{tab:act_trend}. ACT and ACT + Trend are each trained jointly across these six tasks for 10,000 steps. This six-task training setting differs from the official benchmark ACT comparison retained in Table~\ref{tab:robotwin_category}, so the per-task ACT values in the two tables are not directly comparable. Both variants use the same mixed-training and evaluation protocol. The movement trend is predicted from the current observation and supplied as an additional condition; no ground-truth future state is used at test time.

\begin{table}[t]
\caption{ACT with and without movement-trend conditioning in a six-task RoboTwin2.0 validation. Both policies are trained jointly on the six listed tasks for 10,000 steps (success rate \%).}
\label{tab:act_trend}
\centering
\scriptsize
\setlength{\tabcolsep}{4pt}
\begin{tabular}{@{}lccc@{}}
\toprule
Task & ACT & ACT + Trend & Gain \\
\midrule
handover\_block & 42 & 49 & $+7$ \\
place\_cans\_plasticbox & 16 & 48 & $+32$ \\
stack\_bowls\_two & 25 & 73 & $+48$ \\
move\_can\_pot & 22 & 67 & $+45$ \\
click\_alarmclock & 32 & 63 & $+31$ \\
hanging\_mug & 7 & 10 & $+3$ \\
\midrule
\textbf{Mean} & 24.00 & \bestcell{51.67} & $+27.67$ \\
\bottomrule
\end{tabular}
\end{table}

In this six-task validation, movement-trend conditioning improves ACT on all evaluated tasks after 10,000 training steps. The mean success rate increases from 24.00\% to 51.67\%. This result suggests that the learned trend can provide useful guidance beyond the DP3 diffusion backbone. We limit this observation to the evaluated ACT setting and do not claim that the same gain holds for every policy architecture.

\section{Conclusion}

We presented Movement Trend Guidance for 3D diffusion policy learning, addressing the gap between dense action generation and implicit future interaction intent. Our method extracts a compact latent trend representation from a short history of point-cloud and robot-state observations and uses it as soft global conditioning. Because the latent is not decoded into a waypoint sequence or tracking target, the action policy can use future information while retaining the freedom to correct local motion from current geometry. The latent is available through the standard global-conditioning pathway, while an additional gated FiLM branch is restricted to the compressed UNet bottleneck. The trend encoder and action diffusion policy are optimized jointly with end-to-end action diffusion training and an auxiliary future gripper-target signal; a parameter-matched ablation removes this signal while retaining the same latent architecture. Experiments across RoboTwin2.0, LIBERO-40, DexArt, and five SO101 real-robot tasks demonstrate the value of this compact guidance mechanism. The real-robot pipeline obtains calibrated point clouds from Intel RealSense RGB-D observations. Overall, latent movement trends provide a lightweight way to add foresight to 3D diffusion policies without turning future information into a hard plan.

\bibliography{reference}
\bibliographystyle{IEEEtran}

\appendices
\section{Full Per-Task Results on RoboTwin2.0}
\label{app:full_results}

The numerical table in this appendix reports the supplied task-wise results for $\mathit{DP3}^{*}$ and $\mathit{Ours}^{*}$, each trained jointly on all 50 RoboTwin2.0 tasks.

Table~\ref{tab:robotwin_full} reports the success rate (\%) of all methods on each of the 50 RoboTwin2.0 tasks. The starred columns are the 50-task mixed-training results; the other columns retain the official benchmark comparisons.
Tasks are grouped by category.
Category averages and the overall average are included.
The best result in each row is shown in \textbf{bold} with blue shading.

\begin{table*}[h]
\caption{%
  \textbf{Per-task success rates (\%) on RoboTwin2.0 (50 tasks).}
  The starred columns report the 50-task mixed-training systems with the same learned 64-dimensional task embedding: $\mathit{DP3}^{*}$ otherwise follows vanilla DP3, while $\mathit{Ours}^{*}$ adds the proposed movement-trend pathway and bottleneck modulation. The other columns retain the official benchmark comparisons. The best result in each row is in \textbf{bold} with blue shading.
}
\label{tab:robotwin_full}
\centering
\scriptsize
\setlength{\tabcolsep}{4pt}
\begin{tabular}{@{}lcccccc@{}}
\toprule
\textbf{Task} & RDT & $\pi_0$ & ACT & DP & $\mathit{DP3}^{*}$ & $\mathit{Ours}^{*}$ \\
\midrule
\multicolumn{7}{l}{\textit{Single-Arm Object Handling (7 tasks)}} \\
adjust\_bottle              & 81 & 90 & 97 & 97 & 95 & \bestcell{100} \\
grab\_roller                & 74 & 96 & 94 & 98 & 82 & \bestcell{100} \\
shake\_bottle               & 84 & 99 & 63 & 59 & 98 & \bestcell{100} \\
shake\_bottle\_horizontally  & 74 & 97 & 74 & 65 & 98 & \bestcell{100} \\
lift\_pot                   & 72 & 84 & 88 & 39 & \bestcell{100} & 94 \\
dump\_bin\_bigbin            & 64 & 83 & 68 & 49 & 66 & \bestcell{90} \\
hanging\_mug                & 23 & 11 & 7 & 8 & 20 & \bestcell{30} \\
\rowcolor{gray!10}
\textit{Category Avg.}      & 67.4 & 80.0 & 70.1 & 59.3 & 79.9 & \bestcell{87.7} \\
\midrule
\multicolumn{7}{l}{\textit{Articulated Object Interaction (7 tasks)}} \\
click\_alarmclock           & 61 & 63 & 32 & 61 & \bestcell{93} & \bestcell{93} \\
click\_bell                 & 80 & 44 & 58 & 54 & 96 & \bestcell{99} \\
open\_laptop                & 59 & \bestcell{85} & 56 & 49 & 76 & 83 \\
open\_microwave             & 37 & 80 & \bestcell{86} & 5 & 54 & 64 \\
turn\_switch                & 35 & 27 & 5 & 36 & 60 & \bestcell{66} \\
press\_stapler              & 41 & 62 & 31 & 6 & \bestcell{88} & 78 \\
rotate\_QRcode              & 50 & 68 & 1 & 13 & 40 & \bestcell{77} \\
\rowcolor{gray!10}
\textit{Category Avg.}      & 51.9 & 61.3 & 38.4 & 32.0 & 72.4 & \bestcell{80.0} \\
\midrule
\multicolumn{7}{l}{\textit{Bimanual Handover (2 tasks)}} \\
handover\_block             & 45 & 45 & 42 & 10 & 76 & \bestcell{88} \\
handover\_mic               & 90 & 98 & 85 & 53 & 95 & \bestcell{100} \\
\rowcolor{gray!10}
\textit{Category Avg.}      & 67.5 & 71.5 & 63.5 & 31.5 & 85.5 & \bestcell{94.0} \\
\midrule
\multicolumn{7}{l}{\textit{Pick \& Transport (8 tasks)}} \\
move\_can\_pot              & 25 & 58 & 22 & 39 & 85 & \bestcell{86} \\
move\_pillbottle\_pad        & 8 & 21 & 0 & 1 & 22 & \bestcell{44} \\
move\_playingcard\_away      & 43 & 53 & 36 & 47 & 46 & \bestcell{67} \\
move\_stapler\_pad           & 2 & 0 & 0 & 1 & 2 & \bestcell{13} \\
pick\_diverse\_bottles       & 2 & 27 & 7 & 6 & \bestcell{79} & 78 \\
pick\_dual\_bottles          & 42 & 57 & 31 & 24 & 78 & \bestcell{84} \\
put\_bottles\_dustbin        & 21 & 54 & 27 & 22 & 59 & \bestcell{63} \\
put\_object\_cabinet         & 33 & 68 & 15 & 42 & \bestcell{74} & 68 \\
\rowcolor{gray!10}
\textit{Category Avg.}      & 22.0 & 42.3 & 17.3 & 22.8 & 55.6 & \bestcell{62.9} \\
\midrule
\multicolumn{7}{l}{\textit{Precise Placement (17 tasks)}} \\
place\_a2b\_left             & 3 & 31 & 1 & 2 & 30 & \bestcell{48} \\
place\_a2b\_right            & 1 & 27 & 0 & 13 & 36 & \bestcell{55} \\
place\_bread\_basket         & 10 & 17 & 6 & 14 & 53 & \bestcell{54} \\
place\_bread\_skillet        & 5 & 23 & 7 & 11 & 50 & \bestcell{55} \\
place\_burger\_fries         & 50 & 80 & 49 & 72 & \bestcell{81} & 66 \\
place\_can\_basket           & 19 & 41 & 1 & 18 & 80 & \bestcell{81} \\
place\_cans\_plasticbox      & 6 & 34 & 16 & 40 & \bestcell{99} & 96 \\
place\_container\_plate      & 78 & 88 & 72 & 41 & \bestcell{93} & 91 \\
place\_dual\_shoes           & 4 & 15 & 9 & 8 & \bestcell{20} & 16 \\
place\_empty\_cup            & 56 & 37 & 61 & 37 & 66 & \bestcell{91} \\
place\_fan                  & 12 & 20 & 1 & 3 & 18 & \bestcell{24} \\
place\_mouse\_pad            & 1 & \bestcell{7} & 0 & 0 & 1 & 5 \\
place\_object\_basket        & 33 & 16 & 15 & 15 & 43 & \bestcell{59} \\
place\_object\_scale         & 1 & 10 & 0 & 1 & 11 & \bestcell{13} \\
place\_object\_stand         & 15 & 36 & 1 & 22 & 60 & \bestcell{65} \\
place\_phone\_stand          & 15 & 35 & 2 & 13 & \bestcell{62} & 52 \\
place\_shoe                 & 35 & 28 & 5 & 23 & 35 & \bestcell{61} \\
\rowcolor{gray!10}
\textit{Category Avg.}      & 20.2 & 32.1 & 14.5 & 19.6 & 49.3 & \bestcell{54.8} \\
\midrule
\multicolumn{7}{l}{\textit{Multi-Object Stacking (4 tasks)}} \\
stack\_blocks\_two           & 21 & \bestcell{42} & 25 & 7 & 27 & 35 \\
stack\_blocks\_three         & 2 & \bestcell{17} & 0 & 0 & 1 & 3 \\
stack\_bowls\_two            & 76 & \bestcell{91} & 82 & 61 & 73 & 88 \\
stack\_bowls\_three          & 51 & 66 & 48 & 63 & 50 & \bestcell{67} \\
\rowcolor{gray!10}
\textit{Category Avg.}      & 37.5 & \bestcell{54.0} & 38.8 & 32.8 & 37.8 & 48.3 \\
\midrule
\multicolumn{7}{l}{\textit{Tool Use \& Fine-Grained Precision (5 tasks)}} \\
beat\_block\_hammer          & 77 & 43 & 56 & 42 & 70 & \bestcell{80} \\
stamp\_seal                 & 1 & 3 & 2 & 2 & \bestcell{30} & \bestcell{30} \\
scan\_object                & 4 & 18 & 2 & 9 & \bestcell{31} & 28 \\
blocks\_Ranking\_Size        & 3 & \bestcell{19} & 1 & 0 & 3 & 6 \\
blocks\_Ranking\_RGB         & 0 & \bestcell{7} & 0 & 1 & 2 & 5 \\
\rowcolor{gray!10}
\textit{Category Avg.}      & 17.0 & 18.0 & 12.2 & 10.8 & 27.2 & \bestcell{29.8} \\
\midrule
\rowcolor{gray!20}
\textbf{Overall Average}    & 34.5 & 46.4 & 29.7 & 28.0 & 56.1 & \bestcell{62.8} \\
\bottomrule
\end{tabular}
\end{table*}

\section{Experimental Details and Additional RoboTwin2.0 Analysis}
\label{app:experimental_details}

RoboTwin2.0~\cite{chen2025robotwin} contains 50 bimanual manipulation tasks organized around single-arm object handling, articulated object interaction, bimanual handover, pick-and-transport, precise placement, multi-object stacking, and tool use with fine-grained precision.
Each task uses 50 expert demonstrations for training and is evaluated over 100 episodes. We separately and strictly reproduce the official RoboTwin2.0 setting, and additionally evaluate a 50-task mixed-training setting to assess the model's multi-task capability.
On RoboTwin2.0, we compare against five representative methods spanning different policy families: RDT~\cite{liu2024rdt}, $\pi_0$~\cite{black2024pi0}, ACT~\cite{zhao2023act}, Diffusion Policy (DP)~\cite{chi2023diffusionpolicy}, and DP3~\cite{ze2024dp3}.
All baseline numbers are taken from the official RoboTwin2.0 benchmark evaluation.
For our method, the movement-trend encoder and action diffusion policy are optimized jointly throughout end-to-end training.
The full latent-guidance implementation uses an auxiliary future gripper-target loss; the parameter-matched LIBERO ablation removes this loss while keeping the latent architecture fixed.
The latent predictor uses a 256-dimensional hidden representation. Both starred policies receive the same learned 64-dimensional task embedding, which is projected into the observation features before global conditioning of the action diffusion policy. $\mathit{DP3}^{*}$ otherwise follows vanilla DP3, while $\mathit{Ours}^{*}$ adds the trend pathway and bottleneck modulation; controlled ablations retain the same task embedding for component attribution.
Both the trend encoder and action diffusion policy use AdamW with a learning rate of $1.0 \times 10^{-4}$.

All simulation data used in our experiments is generated from the corresponding benchmark simulators rather than collected from physical robots.
Specifically, RoboTwin2.0 demonstrations and evaluation rollouts are generated with the RoboTwin2.0 simulator, and the DexArt experiments are generated with the DexArt simulator.
For real-robot experiments, we manually collect demonstrations on the SO101 dual-arm platform through teleoperation, as described below.

For DexArt, each task uses 100 demonstrations.
Policies are trained for 3000 epochs, evaluated every 200 epochs over 20 episodes, and reported by averaging the best five evaluation checkpoints. This benchmark-specific aggregation differs from the fixed epoch-1000 LIBERO report.

\begin{figure*}[t]
\centering
\includegraphics[width=0.8\textwidth]{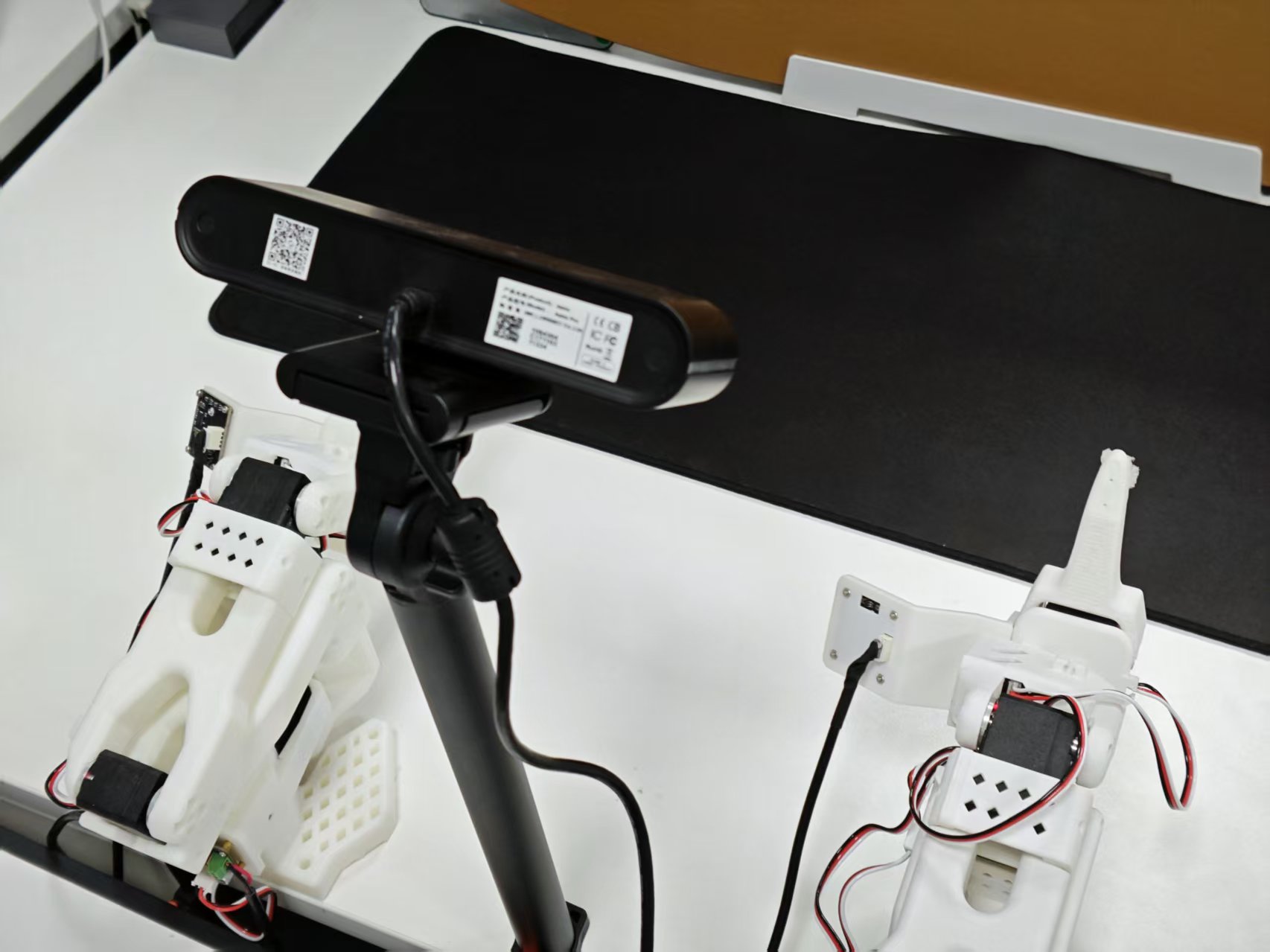}
\caption{Real-robot experimental setup with the SO101 dual-arm platform and Intel RealSense RGB-D cameras. Stereo depth is deprojected with camera intrinsics and transformed with calibrated camera-to-robot extrinsics to obtain the policy point cloud.}
\label{fig:real_setup}
\end{figure*}

For real-robot evaluation, we use the open-source SO101 dual-arm platform on push cube, stack bowls, stack cubes, lift basket, and handover bottle.
The platform is observed with Intel RealSense RGB-D cameras. Each camera provides a synchronized RGB stream and stereo-depth stream; the depth image is deprojected with the calibrated camera intrinsics and transformed into the policy coordinate frame with the camera-to-robot extrinsics.
After invalid-depth filtering and point sampling, the resulting calibrated point cloud is used as the same 3D observation modality as in simulation.
The corresponding real-robot experimental setup is shown in Figure~\ref{fig:real_setup}.
We collect 50 real-robot demonstrations for each task using LeRobot teleoperation and evaluate each task over 20 rollouts.

\begin{figure*}[t]
\centering
\includegraphics[width=\textwidth]{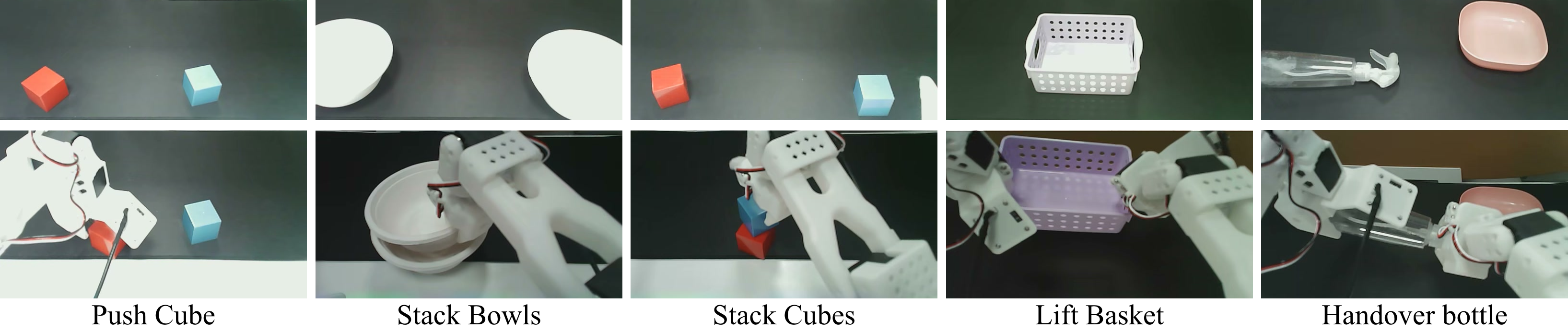}
\caption{Real-robot demonstrations on five manipulation tasks.}
\label{fig:real_robot}
\end{figure*}

The task-level trends in Table~\ref{tab:robotwin_full} align with the intended role of movement trend guidance.
Bimanual handover improves by +8.5 points, where the two arms must coordinate approach, grasp, transfer, and release.
Latent trend encoding captures future interaction intent for the arms, providing soft coordination cues that are not explicit in the observation window alone.
Multi-object stacking shows the largest category-level gain (+10.5), since stacking requires target height and ordering information several steps ahead.
Single-arm object handling (+7.9), articulated object interaction (+7.6), and pick-and-transport (+7.3) similarly involve multi-phase behavior where compressed future information can guide smoother action generation.
The gains are smaller on precise placement (+5.5) and tool-use tasks (+2.6), where success depends strongly on fine-grained local geometry.

We further partition the 50 RoboTwin2.0 tasks by $\mathit{DP3}^{*}$ baseline success rate.
Among the 11 tasks with $\mathit{DP3}^{*}$ success rate below 30\%, our method improves on 10 and achieves an average gain of +6.1 points.
For the five tasks with $\mathit{DP3}^{*}$ success rates from 30\% to below 40\%, the average gain is +12.0 points.
Among the 21 tasks with $\mathit{DP3}^{*}$ success rates from 40\% to 80\%, the average gain is +9.8 points, where movement trend guidance provides the most complementary value to the existing geometric perception.
Among the 13 tasks with $\mathit{DP3}^{*}$ success rates above 80\%, the average change is 0.0 points, reflecting a near-ceiling effect and broadly comparable performance overall.
This pattern suggests that movement trend guidance is most beneficial when the base policy has learned a reasonable action prior but lacks explicit geometric cues about the future movement trend.

\section{Complete LIBERO-40 Checkpoint Results}
\label{app:libero_full_results}
Tables~\ref{tab:libero_dp3_checkpoints} and \ref{tab:libero_ours_checkpoints} report every saved checkpoint in the LIBERO-40 evaluation. Each entry is the suite-level success rate (\%) written as mean $\pm$ sample standard deviation across rollout seeds 7, 17, and 27. Each suite contains 10 tasks and 1,500 evaluation episodes per checkpoint; Overall contains all 40 tasks and 6,000 episodes.
All values are computed directly from the bounded seed-level rates in the accompanying CSV.

\begin{table*}[p]
\caption{DP3 LIBERO-40 success rates across all saved checkpoints (mean $\pm$ sample standard deviation across three rollout seeds, \%).}
\label{tab:libero_dp3_checkpoints}
\centering\small
\setlength{\tabcolsep}{8pt}
\begin{tabular}{@{}lccccc@{}}
\toprule
Epoch & LIBERO-Spatial & LIBERO-Object & LIBERO-Goal & LIBERO-10 & Overall \\
\midrule
100 & 0.93 $\pm$ 0.46 & 3.73 $\pm$ 0.23 & 1.67 $\pm$ 1.15 & 1.20 $\pm$ 0.20 & 1.88 $\pm$ 0.24 \\
200 & 3.40 $\pm$ 0.20 & 5.80 $\pm$ 0.20 & 1.73 $\pm$ 0.31 & 0.33 $\pm$ 0.31 & 2.82 $\pm$ 0.10 \\
300 & 7.67 $\pm$ 1.17 & 12.47 $\pm$ 2.10 & 1.87 $\pm$ 0.90 & 4.67 $\pm$ 1.03 & 6.67 $\pm$ 0.70 \\
400 & 6.87 $\pm$ 1.27 & 7.80 $\pm$ 0.87 & 6.40 $\pm$ 0.72 & 9.87 $\pm$ 1.55 & 7.73 $\pm$ 0.70 \\
500 & 8.53 $\pm$ 0.81 & 33.80 $\pm$ 1.06 & 5.87 $\pm$ 0.50 & 18.40 $\pm$ 1.64 & 16.65 $\pm$ 0.35 \\
600 & 20.73 $\pm$ 1.47 & 33.13 $\pm$ 2.32 & 7.33 $\pm$ 0.83 & 28.93 $\pm$ 2.05 & 22.53 $\pm$ 0.23 \\
700 & 18.93 $\pm$ 1.70 & 46.33 $\pm$ 3.86 & 8.53 $\pm$ 0.58 & 31.33 $\pm$ 3.45 & 26.28 $\pm$ 0.81 \\
800 & 19.33 $\pm$ 2.12 & 60.40 $\pm$ 0.20 & 7.87 $\pm$ 0.61 & 42.00 $\pm$ 0.20 & 32.40 $\pm$ 0.43 \\
900 & 26.93 $\pm$ 1.72 & 66.13 $\pm$ 0.81 & 9.93 $\pm$ 1.55 & 41.47 $\pm$ 0.12 & 36.12 $\pm$ 0.15 \\
1000 & 26.40 $\pm$ 1.31 & 67.53 $\pm$ 1.14 & 8.73 $\pm$ 0.76 & 45.67 $\pm$ 1.80 & 37.08 $\pm$ 1.21 \\
\bottomrule
\end{tabular}
\end{table*}

\begin{table*}[p]
\caption{Ours LIBERO-40 success rates across all saved checkpoints (mean $\pm$ sample standard deviation across three rollout seeds, \%).}
\label{tab:libero_ours_checkpoints}
\centering\small
\setlength{\tabcolsep}{8pt}
\begin{tabular}{@{}lccccc@{}}
\toprule
Epoch & LIBERO-Spatial & LIBERO-Object & LIBERO-Goal & LIBERO-10 & Overall \\
\midrule
100 & 35.73 $\pm$ 1.50 & 12.67 $\pm$ 1.14 & 39.27 $\pm$ 1.97 & 8.67 $\pm$ 0.23 & 24.08 $\pm$ 0.74 \\
200 & 22.33 $\pm$ 1.47 & 57.00 $\pm$ 2.12 & 46.07 $\pm$ 1.21 & 17.87 $\pm$ 1.47 & 35.82 $\pm$ 0.80 \\
300 & 65.33 $\pm$ 2.77 & 88.53 $\pm$ 0.76 & 75.40 $\pm$ 3.29 & 48.93 $\pm$ 1.67 & 69.55 $\pm$ 1.49 \\
400 & 57.00 $\pm$ 2.11 & 93.53 $\pm$ 1.21 & 74.53 $\pm$ 1.29 & 60.33 $\pm$ 0.99 & 71.35 $\pm$ 0.74 \\
500 & 62.53 $\pm$ 3.71 & 94.20 $\pm$ 0.80 & 76.07 $\pm$ 2.23 & 57.20 $\pm$ 1.06 & 72.50 $\pm$ 1.19 \\
600 & 60.13 $\pm$ 1.81 & 94.60 $\pm$ 0.60 & 76.13 $\pm$ 1.67 & 59.80 $\pm$ 0.72 & 72.67 $\pm$ 0.28 \\
700 & 57.80 $\pm$ 0.87 & 94.13 $\pm$ 1.01 & 78.27 $\pm$ 1.86 & 57.87 $\pm$ 1.03 & 72.02 $\pm$ 0.29 \\
800 & 59.07 $\pm$ 0.46 & 93.20 $\pm$ 1.20 & 78.40 $\pm$ 0.35 & 57.93 $\pm$ 1.50 & 72.15 $\pm$ 0.61 \\
900 & 58.47 $\pm$ 0.23 & 94.47 $\pm$ 1.10 & 76.80 $\pm$ 1.04 & 59.07 $\pm$ 2.34 & 72.20 $\pm$ 0.69 \\
1000 & 58.20 $\pm$ 0.72 & 93.53 $\pm$ 0.70 & 77.53 $\pm$ 2.32 & 58.47 $\pm$ 0.64 & 71.93 $\pm$ 0.49 \\
\bottomrule
\end{tabular}
\end{table*}

\section{Detailed Ablation Analysis}
\label{app:ablation_analysis}

The main paper summarizes the ablation results with Fig.~\ref{fig:ablation}. Here we provide additional interpretation for each group of variants.

For guidance construction and training, removing gated FiLM reduces the model to a DP3-style policy in which the latent trend is available only through early fusion. The comparison tests whether the latent contributes more than an increase in representation capacity. We also compare parameter-matched variants that replace the default MLP trend predictor with a diffusion predictor or remove input normalization. These variants test whether the result depends on the specific predictor and preprocessing choices. The end-to-end configuration allows the latent to adapt to the action objective while remaining a soft condition that does not prescribe the executed motion.

For future-target offsets, the observation history is fixed to three frames while the supervised offsets are varied. Short-range offsets can be redundant with the observation window, while long-range offsets provide broader movement information. The selected offsets summarize near-term contact and longer-range movement tendency without creating an intermediate target for the policy to reach.

For injection architecture, applying gated FiLM at every UNet block reduces success from the $\mathit{Ours}^{*}$ value of 62.8\% to 55.8\%. This suggests that pervasive gated modulation can make the high-level latent interfere with local action refinement, especially in shallow and upsampling blocks responsible for short-term correction. Replacing gated FiLM with cross-attention performs even worse, reaching 52.9\% and falling below the 56.1\% $\mathit{DP3}^{*}$ baseline. Cross-attention creates a stronger coupling between denoising features and guidance tokens, while gated FiLM uses element-wise affine modulation with a learnable gate initialized near identity. The ordering among bottleneck-only gated FiLM, all-block gated FiLM, and cross-attention supports the conservative injection design: latent trend guidance is most useful when it shapes global action structure without dominating local denoising behavior.

\section{Full Per-Task Ablation Results on RoboTwin2.0}
\label{app:full_ablation_results}

Tables~\ref{tab:ablation_guidance_full}, \ref{tab:ablation_offsets_full}, and \ref{tab:ablation_injection_full} report the complete per-task RoboTwin2.0 ablation results summarized in Figure~\ref{fig:ablation}.

\begin{table*}[p]
\caption{Per-task slices of the 50-task mixed-training ablation for latent guidance construction and training on RoboTwin2.0 (success rate \%). All columns are variants of a shared policy trained jointly on the full suite. The diffusion-predictor column replaces the default MLP trend predictor, matching Fig.~\ref{fig:ablation}.}
\label{tab:ablation_guidance_full}
\centering
\scriptsize
\setlength{\tabcolsep}{3pt}
\begin{tabular}{@{}lccccc@{}}
\toprule
Task & $\mathit{DP3}^{*}$ & $\mathit{Ours}^{*}$ & w/o gate & Diff. pred. & w/o norm. \\
\midrule
\multicolumn{6}{l}{\textit{Single-Arm Object Handling}} \\
adjust\_bottle & 95 & 100 & 100 & 100 & 100 \\
grab\_roller & 82 & 100 & 100 & 99 & 100 \\
shake\_bottle & 98 & 100 & 100 & 98 & 100 \\
shake\_bottle\_horizontally & 98 & 100 & 100 & 99 & 100 \\
lift\_pot & 100 & 94 & 100 & 94 & 97 \\
dump\_bin\_bigbin & 66 & 90 & 89 & 86 & 89 \\
hanging\_mug & 20 & 30 & 21 & 33 & 19 \\
\midrule
\multicolumn{6}{l}{\textit{Articulated Object Interaction}} \\
click\_alarmclock & 93 & 93 & 91 & 88 & 93 \\
click\_bell & 96 & 99 & 100 & 99 & 99 \\
open\_laptop & 76 & 83 & 77 & 75 & 83 \\
open\_microwave & 54 & 64 & 68 & 94 & 67 \\
turn\_switch & 60 & 66 & 47 & 56 & 55 \\
press\_stapler & 88 & 78 & 65 & 75 & 72 \\
rotate\_QRcode & 40 & 77 & 52 & 62 & 61 \\
\midrule
\multicolumn{6}{l}{\textit{Bimanual Handover}} \\
handover\_block & 76 & 88 & 93 & 90 & 94 \\
handover\_mic & 95 & 100 & 100 & 97 & 100 \\
\midrule
\multicolumn{6}{l}{\textit{Pick \& Transport}} \\
move\_can\_pot & 85 & 86 & 79 & 79 & 85 \\
move\_pillbottle\_pad & 22 & 44 & 43 & 35 & 47 \\
move\_playingcard\_away & 46 & 67 & 68 & 62 & 73 \\
move\_stapler\_pad & 2 & 13 & 9 & 14 & 8 \\
pick\_diverse\_bottles & 79 & 78 & 62 & 59 & 65 \\
pick\_dual\_bottles & 78 & 84 & 65 & 53 & 54 \\
put\_bottles\_dustbin & 59 & 63 & 67 & 53 & 68 \\
put\_object\_cabinet & 74 & 68 & 56 & 60 & 70 \\
\midrule
\multicolumn{6}{l}{\textit{Precise Placement}} \\
place\_a2b\_left & 30 & 48 & 26 & 31 & 38 \\
place\_a2b\_right & 36 & 55 & 21 & 41 & 40 \\
place\_bread\_basket & 53 & 54 & 28 & 27 & 29 \\
place\_bread\_skillet & 50 & 55 & 29 & 29 & 30 \\
place\_burger\_fries & 81 & 66 & 65 & 63 & 77 \\
place\_can\_basket & 80 & 81 & 74 & 68 & 78 \\
place\_cans\_plasticbox & 99 & 96 & 69 & 70 & 62 \\
place\_container\_plate & 93 & 91 & 92 & 90 & 86 \\
place\_dual\_shoes & 20 & 16 & 16 & 17 & 12 \\
place\_empty\_cup & 66 & 91 & 90 & 78 & 83 \\
place\_fan & 18 & 24 & 42 & 33 & 39 \\
place\_mouse\_pad & 1 & 5 & 6 & 4 & 6 \\
place\_object\_basket & 43 & 59 & 62 & 56 & 65 \\
place\_object\_scale & 11 & 13 & 16 & 14 & 13 \\
place\_object\_stand & 60 & 65 & 50 & 51 & 53 \\
place\_phone\_stand & 62 & 52 & 41 & 57 & 52 \\
place\_shoe & 35 & 61 & 51 & 54 & 46 \\
\midrule
\multicolumn{6}{l}{\textit{Multi-Object Stacking}} \\
stack\_blocks\_two & 27 & 35 & 25 & 27 & 33 \\
stack\_blocks\_three & 1 & 3 & 3 & 4 & 2 \\
stack\_bowls\_two & 73 & 88 & 92 & 93 & 91 \\
stack\_bowls\_three & 50 & 67 & 79 & 72 & 75 \\
\midrule
\multicolumn{6}{l}{\textit{Tool Use \& Fine-Grained Precision}} \\
beat\_block\_hammer & 70 & 80 & 74 & 68 & 76 \\
stamp\_seal & 30 & 30 & 25 & 29 & 21 \\
scan\_object & 31 & 28 & 24 & 18 & 18 \\
blocks\_Ranking\_Size & 3 & 6 & 3 & 3 & 3 \\
blocks\_Ranking\_RGB & 2 & 5 & 1 & 7 & 2 \\
\midrule
\rowcolor{gray!20}
Average & 56.1 & 62.8 & 57.1 & 57.3 & 58.6 \\
\bottomrule
\end{tabular}
\end{table*}

\begin{table*}[p]
\caption{Per-task slices of the 50-task mixed-training ablation for future-target offset configurations on RoboTwin2.0 (success rate \%). The observation history is fixed to three frames; column names match Fig.~\ref{fig:ablation}. All columns are variants of a shared policy trained jointly on the full suite.}
\label{tab:ablation_offsets_full}
\centering
\scriptsize
\setlength{\tabcolsep}{3pt}
\begin{tabular}{@{}lcccccccc@{}}
\toprule
Task & $\mathit{DP3}^{*}$ & $\mathit{Ours}^{*}$ & 7 offsets & short range & long range & 2 offsets & single near & single far \\
\midrule
\multicolumn{9}{l}{\textit{Single-Arm Object Handling}} \\
adjust\_bottle & 95 & 100 & 100 & 100 & 85 & 99 & 100 & 100 \\
grab\_roller & 82 & 100 & 98 & 99 & 100 & 100 & 100 & 100 \\
shake\_bottle & 98 & 100 & 99 & 96 & 96 & 96 & 98 & 99 \\
shake\_bottle\_horizontally & 98 & 100 & 97 & 98 & 100 & 99 & 98 & 98 \\
lift\_pot & 100 & 94 & 90 & 91 & 96 & 91 & 91 & 92 \\
dump\_bin\_bigbin & 66 & 90 & 84 & 89 & 91 & 86 & 82 & 81 \\
hanging\_mug & 20 & 30 & 21 & 19 & 19 & 26 & 30 & 26 \\
\midrule
\multicolumn{9}{l}{\textit{Articulated Object Interaction}} \\
click\_alarmclock & 93 & 93 & 80 & 78 & 94 & 88 & 84 & 80 \\
click\_bell & 96 & 99 & 100 & 99 & 97 & 99 & 100 & 99 \\
open\_laptop & 76 & 83 & 83 & 83 & 82 & 87 & 80 & 81 \\
open\_microwave & 54 & 64 & 58 & 51 & 59 & 63 & 60 & 85 \\
turn\_switch & 60 & 66 & 55 & 53 & 52 & 59 & 51 & 54 \\
press\_stapler & 88 & 78 & 77 & 74 & 75 & 73 & 71 & 72 \\
rotate\_QRcode & 40 & 77 & 50 & 74 & 52 & 57 & 55 & 74 \\
\midrule
\multicolumn{9}{l}{\textit{Bimanual Handover}} \\
handover\_block & 76 & 88 & 92 & 85 & 89 & 83 & 91 & 87 \\
handover\_mic & 95 & 100 & 91 & 97 & 66 & 97 & 99 & 100 \\
\midrule
\multicolumn{9}{l}{\textit{Pick \& Transport}} \\
move\_can\_pot & 85 & 86 & 88 & 86 & 86 & 81 & 81 & 83 \\
move\_pillbottle\_pad & 22 & 44 & 33 & 37 & 40 & 34 & 34 & 24 \\
move\_playingcard\_away & 46 & 67 & 71 & 73 & 83 & 70 & 74 & 71 \\
move\_stapler\_pad & 2 & 13 & 17 & 10 & 4 & 12 & 4 & 8 \\
pick\_diverse\_bottles & 79 & 78 & 54 & 52 & 68 & 62 & 54 & 55 \\
pick\_dual\_bottles & 78 & 84 & 56 & 54 & 73 & 67 & 70 & 57 \\
put\_bottles\_dustbin & 59 & 63 & 62 & 61 & 72 & 58 & 54 & 49 \\
put\_object\_cabinet & 74 & 68 & 68 & 65 & 68 & 59 & 68 & 57 \\
\midrule
\multicolumn{9}{l}{\textit{Precise Placement}} \\
place\_a2b\_left & 30 & 48 & 30 & 27 & 39 & 31 & 30 & 26 \\
place\_a2b\_right & 36 & 55 & 40 & 38 & 41 & 34 & 40 & 38 \\
place\_bread\_basket & 53 & 54 & 32 & 20 & 28 & 16 & 18 & 22 \\
place\_bread\_skillet & 50 & 55 & 35 & 30 & 31 & 31 & 33 & 27 \\
place\_burger\_fries & 81 & 66 & 70 & 72 & 70 & 68 & 74 & 73 \\
place\_can\_basket & 80 & 81 & 76 & 73 & 65 & 73 & 74 & 76 \\
place\_cans\_plasticbox & 99 & 96 & 84 & 56 & 69 & 69 & 56 & 61 \\
place\_container\_plate & 93 & 91 & 80 & 86 & 91 & 85 & 83 & 87 \\
place\_dual\_shoes & 20 & 16 & 14 & 10 & 13 & 17 & 17 & 11 \\
place\_empty\_cup & 66 & 91 & 80 & 81 & 86 & 81 & 76 & 69 \\
place\_fan & 18 & 24 & 37 & 39 & 41 & 45 & 42 & 33 \\
place\_mouse\_pad & 1 & 5 & 2 & 3 & 5 & 1 & 4 & 3 \\
place\_object\_basket & 43 & 59 & 60 & 54 & 64 & 50 & 48 & 52 \\
place\_object\_scale & 11 & 13 & 8 & 13 & 11 & 7 & 9 & 6 \\
place\_object\_stand & 60 & 65 & 61 & 50 & 55 & 49 & 51 & 52 \\
place\_phone\_stand & 62 & 52 & 56 & 42 & 49 & 43 & 39 & 48 \\
place\_shoe & 35 & 61 & 48 & 46 & 59 & 51 & 45 & 51 \\
\midrule
\multicolumn{9}{l}{\textit{Multi-Object Stacking}} \\
stack\_blocks\_two & 27 & 35 & 24 & 27 & 24 & 24 & 31 & 29 \\
stack\_blocks\_three & 1 & 3 & 1 & 2 & 2 & 0 & 1 & 2 \\
stack\_bowls\_two & 73 & 88 & 90 & 92 & 90 & 80 & 78 & 88 \\
stack\_bowls\_three & 50 & 67 & 60 & 68 & 78 & 68 & 62 & 69 \\
\midrule
\multicolumn{9}{l}{\textit{Tool Use \& Fine-Grained Precision}} \\
beat\_block\_hammer & 70 & 80 & 66 & 66 & 63 & 76 & 61 & 52 \\
stamp\_seal & 30 & 30 & 29 & 29 & 25 & 23 & 22 & 27 \\
scan\_object & 31 & 28 & 19 & 16 & 21 & 19 & 19 & 23 \\
blocks\_Ranking\_Size & 3 & 6 & 3 & 3 & 3 & 1 & 1 & 1 \\
blocks\_Ranking\_RGB & 2 & 5 & 2 & 2 & 2 & 2 & 3 & 1 \\
\midrule
\rowcolor{gray!20}
Average & 56.1 & 62.8 & 56.6 & 55.4 & 57.4 & 55.8 & 54.9 & 55.2 \\
\bottomrule
\end{tabular}
\end{table*}

\begin{table*}[p]
\caption{Per-task slices of the 50-task mixed-training ablation for movement-trend guidance injection on RoboTwin2.0 (success rate \%). All columns are variants of a shared policy trained jointly on the full suite.}
\label{tab:ablation_injection_full}
\centering
\scriptsize
\setlength{\tabcolsep}{3pt}
\begin{tabular}{@{}lcccc@{}}
\toprule
Task & $\mathit{DP3}^{*}$ & $\mathit{Ours}^{*}$ & All-block & Cross-attn \\
\midrule
\multicolumn{5}{l}{\textit{Single-Arm Object Handling}} \\
adjust\_bottle & 95 & 100 & 100 & 0 \\
grab\_roller & 82 & 100 & 99 & 99 \\
shake\_bottle & 98 & 100 & 98 & 95 \\
shake\_bottle\_horizontally & 98 & 100 & 100 & 98 \\
lift\_pot & 100 & 94 & 100 & 93 \\
dump\_bin\_bigbin & 66 & 90 & 80 & 80 \\
hanging\_mug & 20 & 30 & 20 & 23 \\
\midrule
\multicolumn{5}{l}{\textit{Articulated Object Interaction}} \\
click\_alarmclock & 93 & 93 & 79 & 90 \\
click\_bell & 96 & 99 & 99 & 99 \\
open\_laptop & 76 & 83 & 82 & 77 \\
open\_microwave & 54 & 64 & 75 & 71 \\
turn\_switch & 60 & 66 & 53 & 54 \\
press\_stapler & 88 & 78 & 76 & 70 \\
rotate\_QRcode & 40 & 77 & 75 & 76 \\
\midrule
\multicolumn{5}{l}{\textit{Bimanual Handover}} \\
handover\_block & 76 & 88 & 78 & 98 \\
handover\_mic & 95 & 100 & 89 & 96 \\
\midrule
\multicolumn{5}{l}{\textit{Pick \& Transport}} \\
move\_can\_pot & 85 & 86 & 81 & 85 \\
move\_pillbottle\_pad & 22 & 44 & 44 & 22 \\
move\_playingcard\_away & 46 & 67 & 78 & 67 \\
move\_stapler\_pad & 2 & 13 & 8 & 7 \\
pick\_diverse\_bottles & 79 & 78 & 48 & 55 \\
pick\_dual\_bottles & 78 & 84 & 72 & 41 \\
put\_bottles\_dustbin & 59 & 63 & 55 & 60 \\
put\_object\_cabinet & 74 & 68 & 63 & 65 \\
\midrule
\multicolumn{5}{l}{\textit{Precise Placement}} \\
place\_a2b\_left & 30 & 48 & 21 & 35 \\
place\_a2b\_right & 36 & 55 & 41 & 36 \\
place\_bread\_basket & 53 & 54 & 24 & 23 \\
place\_bread\_skillet & 50 & 55 & 31 & 25 \\
place\_burger\_fries & 81 & 66 & 65 & 60 \\
place\_can\_basket & 80 & 81 & 70 & 69 \\
place\_cans\_plasticbox & 99 & 96 & 38 & 52 \\
place\_container\_plate & 93 & 91 & 82 & 86 \\
place\_dual\_shoes & 20 & 16 & 10 & 14 \\
place\_empty\_cup & 66 & 91 & 89 & 84 \\
place\_fan & 18 & 24 & 38 & 42 \\
place\_mouse\_pad & 1 & 5 & 5 & 2 \\
place\_object\_basket & 43 & 59 & 61 & 52 \\
place\_object\_scale & 11 & 13 & 10 & 8 \\
place\_object\_stand & 60 & 65 & 52 & 52 \\
place\_phone\_stand & 62 & 52 & 46 & 48 \\
place\_shoe & 35 & 61 & 63 & 44 \\
\midrule
\multicolumn{5}{l}{\textit{Multi-Object Stacking}} \\
stack\_blocks\_two & 27 & 35 & 22 & 28 \\
stack\_blocks\_three & 1 & 3 & 1 & 2 \\
stack\_bowls\_two & 73 & 88 & 86 & 87 \\
stack\_bowls\_three & 50 & 67 & 61 & 73 \\
\midrule
\multicolumn{5}{l}{\textit{Tool Use \& Fine-Grained Precision}} \\
beat\_block\_hammer & 70 & 80 & 63 & 57 \\
stamp\_seal & 30 & 30 & 21 & 28 \\
scan\_object & 31 & 28 & 29 & 9 \\
blocks\_Ranking\_Size & 3 & 6 & 4 & 3 \\
blocks\_Ranking\_RGB & 2 & 5 & 5 & 7 \\
\midrule
\rowcolor{gray!20}
Average & 56.1 & 62.8 & 55.8 & 52.9 \\
\bottomrule
\end{tabular}
\end{table*}

\end{document}